\documentclass[10pt,letterpaper]{article}
\usepackage{cogsci}
\usepackage{pslatex}
\usepackage{apacite}
\usepackage[pdftex]{graphicx}  
\usepackage[font=small,labelfont=bf]{caption}
\usepackage[font=small,skip=0pt]{caption}
\usepackage{enumitem}
\pdfoutput=1

\title{Representational efficiency outweighs action efficiency \\ in human program induction}
\author{
    {\large \bf Sophia Sanborn\textsuperscript{\scriptsize 1},
     \large \bf David D. Bourgin\textsuperscript{\scriptsize 1},
      \large \bf Michael Chang\textsuperscript{\scriptsize 2},
     \large \bf Thomas L. Griffiths\textsuperscript{\scriptsize 1,2}}\\
     {\large \bf \textsuperscript{\scriptsize 1}} Department of Psychology\\ 
     {\large \bf \textsuperscript{\scriptsize 2}} Department of Electrical Engineering and Computer Science\\ 
     University of California, Berkeley\\
     Berkeley, CA, 94720 USA\\
     \texttt{
        \{sanborn,ddbourgin,mbchang,tom\_griffiths\}@berkeley.edu
     }
}

\begin{document}

\maketitle

\begin{abstract}
The importance of hierarchically structured representations for tractable planning has long been acknowledged. However, the questions of how people discover such abstractions and how to define a set of optimal abstractions remain open. This problem has been explored in cognitive science in the problem solving literature and in computer science in hierarchical reinforcement learning. Here, we emphasize an algorithmic perspective on learning hierarchical representations in which the objective is to efficiently encode the structure of the problem, or, equivalently, to learn an algorithm with minimal length. We introduce a novel problem-solving paradigm that links problem solving and program induction under the Markov Decision Process (MDP) framework. Using this task, we target the question of whether humans discover hierarchical solutions by maximizing efficiency in number of actions they generate or by minimizing the complexity of the resulting representation and find evidence for the primacy of representational efficiency. 

\textbf{Keywords:} 
problem solving; program induction; hierarchical reinforcement learning
\end{abstract}

\section{Introduction}
The space of simple actions available to humans navigating the world is immense. Even a seemingly straightforward task such as turning a doorknob is composed of smaller actions: arranging one's fingers and thumb into a grasping position and rotating one's wrist. Each of these actions is composed of smaller muscular movements, each of which is ultimately controlled by the firing of  neurons in motor cortex. With even a small number of primitive actions, the space of possible finite-length sequences that an individual can execute explodes. Despite this, the human brain manages to select sequences of actions remarkably effectively, constructing and executing plans that lead to the efficient achievement of goals.

The formation and utilization of hierarchical representations dramatically reduces the complexity of this planning problem \cite{simon1991architecture}. This is the strategy that the brain employs; hierarchy is ubiquitous in neural and cognitive representations in nearly every domain, ranging from language, perception, motor control, and problem solving \cite{reason:Chomsky57a, van1983hierarchical, chase1973perception, botvinick2009hierarchically}. The formation of hierarchical representations confers considerable adaptive benefit by supporting efficient coding, reducing working memory demands, and enabling a learner to perceive higher-order structure that would otherwise be too computationally costly to extract.

These insights from human perception and cognition have inspired methods in the field of hierarchical reinforcement learning (HRL), which integrates structured hierarchical representations into the reinforcement learning framework \cite{barto2003recent}. Early work in HRL used hand-designed abstractions, such as sub-goals, to demonstrate that a well-designed hierarchy can significantly speed the discovery of shortest-path solutions \cite{sutton1999between} and enable agents to learn to solve related tasks more quickly \cite{konidaris2007building}. More recent work in HRL has investigated approaches for learning useful hierarchical representations \cite{barto2013behavioral,botvinick2009hierarchically,csimcsek2009skill,vigorito2010intrinsically,van2011grounding}. Across this work, one consistent finding is that small differences in the formulation of the task objective may lead to radically different hierarchical decompositions. A natural question thus becomes: given a particular problem, what defines an ``optimal" hierarchy for action? 

For state-of-the-art hierarchical reinforcement learning algorithms, the utility of a hierarchy is typically assessed by its effect on the agent's learning curve---how greatly the hierarchy accelerates the agent's convergence to the shortest-path solution. In the cognitive science literature, the focus has been on properties of the hierarchical representation itself, and researchers have aimed to determine hierarchies that capture the structure of the task.In some cases, hierarchies optimized to minimize the number of primitive actions used in the solution and hierarchies optimized to capture the structure of the task may map on to each other. However, these objectives can deviate in important ways: the shortest-path solution may not possess the most structured hierarchical representation. 

In this paper, we aim to bridge approaches to hierarchy learning in computer science and cognitive science by introducing a novel problem-solving task that is naturally formulated as a Markov decision process and enables participants to explicitly articulate algorithmically-structured solutions. Ultimately, this task may facilitate direct comparisons between the algorithms used for HRL in computer science and the hierarchical solutions generated by humans, and may be used as a novel benchmark for HRL algorithms. In this paper, we use this task to examine the properties of the hierarchies that humans generate in problem solving. In particular, we target the question of whether humans generate hierarchical solutions that accord with the objective typically optimized in HRL---efficiency in terms of primitive actions---or if humans express a bias to generate hierarchies that capture the structure of the task and yield lower complexity representations.

We focus here on a general measure of representational complexity: the length of the shortest program that generates a sequence of actions. This metric is inspired by Kolmogorov complexity \cite{kolmogorov1963tables} and is in accordance with the principle of efficient coding in neuroscience \cite{barlow1961possible}. Extending the efficient coding hypothesis to higher-level cognition, we hypothesize that humans possess a bias for algorithmic simplicity and coding efficiency that may guide their specification of hierarchical solutions in problem solving. 

Our perspective emphasizes the importance of representation learning in reinforcement learning and problem solving, and imports insights from sensory neuroscience into higher-level cognition. The relevance of this connection has been expressed by several researchers \cite{tishby2011information, botvinick2015reinforcement}, and preliminary empirical support for efficient coding in problem solving has been found in a series of studies that show that humans are sensitive to the latent hierarchical structure in flat solutions and show preferences for maximizing computational efficiency when multiple shortest flat paths are available \cite{solway2014optimal}. Although these findings are suggestive, the structure present in the flat sequences in these studies is limited to ``bottlenecks'' in the state space. Further, the strength of the bias for computational efficiency is unclear since the computationally efficient solutions are also maximally efficient in terms of actions. Here, we design contexts that express richer, algorithmic structure and explicitly tease apart and trade off efficiency of action and representation, finding evidence for the primacy of representational efficiency.

\section{Background}

The theory of Markov decision processes (MDPs; Sutton \& Barto, 1998) provides a computational framework for modeling problem solving as search within a metaphorical {\em problem} or {\em state space}. Optimal solutions to an MDP may be obtained via dynamic programming or reinforcement learning. In an MDP, the problem is formalized as a set of {\em states} linked by {\em actions}. Transitions between states can be stochastic or deterministic, and pairs of states and actions yield variable {\em rewards}. The goal is to find the {\em policy} governing the selection of actions in each state that yields the greatest expected reward. The shortest-path solution to a problem is the policy that obtains the greatest expected reward with the fewest number of steps. 

To integrate hierarchical structure, the problem formulation may be extended to include abstractions over the ``primitive'' actions or states. In the formulation of the task considered in this paper, we introduce hierarchy by distinguishing primitive actions from sub-processes---stored sequences of primitive actions that may be executed with a single call---and we consider deterministic solutions (programs) consisting of calls to primitive actions and sub-processes. Sub-processes may be nested to yield hierarchies of arbitrary depth. We contrast hierarchical solutions to ``flat'' solutions that use primitive actions alone. We further distinguish the flat path of actions that a program generates from the stored representation of the program itself. The flat path generated by a program may or may not map onto the optimal flat solution to a problem, which allows us to tease apart the objectives of action efficiency and representational efficiency.

\section{Lightbot: An experimental paradigm for hierarchical problem solving}

\begin{figure}[ht]
\begin{center}
\includegraphics[width=.4\textwidth]{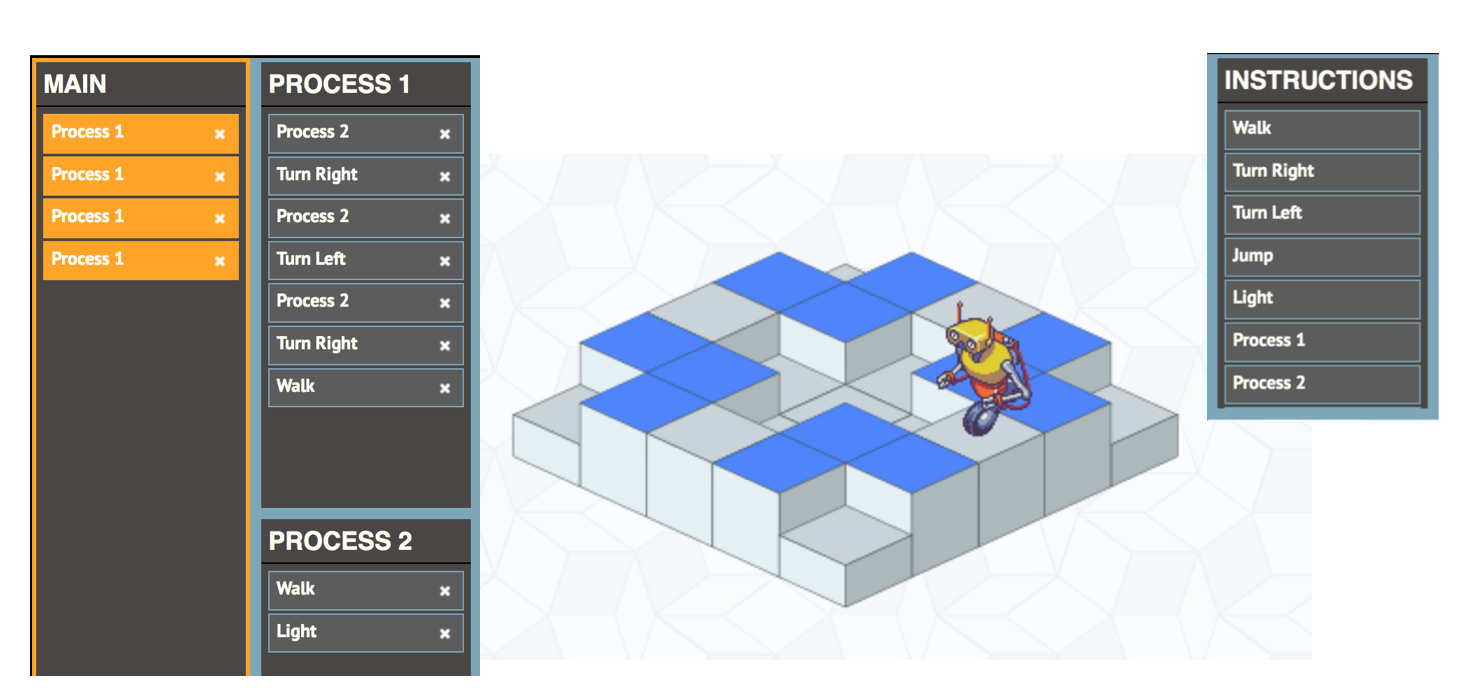}
\end{center}
\caption{Lightbot screenshot showing puzzle and program.} 
\label{fig:lightbot}
\end{figure}

We present a novel problem-solving setting that is rich enough to elicit explicit hierarchical, algorithmically-structured solutions from participants and is amenable to formalization in the RL framework. The task is adapted from a game designed to teach children how to program (Lightbot: \url{https://lightbot.com}; Figure \ref{fig:lightbot})\footnote{The codebase that we adapted was developed by Laurent Haan (\url{https://github.com/haan/Lightbot})}.
 
In the Lightbot domain, players control a robot in a 3-dimensional block world and must determine a sequence of actions that will lead the robot to turn on all of the light tiles (blue tiles) in the world. There are five primitive actions (walk, jump, turn right, turn left, and toggle light). Participants may additionally be given the capacity to store sequences of primitive actions as sub-processes. These processes can be called within the main program, within each other, or within themselves, thus allowing participants to generate hierarchically structured, algorithmic solutions that may incorporate simple recursion.  

Aside from the behavior it elicits, this task has many properties that make it appealing from a methodological perspective: (a) problems of arbitrary complexity and difficulty can be generated, (b) participants explicitly externalize their problem-solving strategies by writing down the algorithm they use to solve the problem, and (c) the process of generating and testing these solutions can be recorded and analyzed. 

The task is also interesting from a reinforcement learning perspective, as it poses an objective function not typically examined in the RL literature. In the Lightbot task, the objective is not simply to maximize reward by turning on as many lights as possible; the objective here is to do so as \textit{computationally efficiently} as possible by generating an algorithm that exploits the abstract structure of the problem. This task provides a bridge for comparing hierarchies generated by humans and by HRL algorithms and may be used as a benchmark for HRL in the future.

\subsection{Action efficiency vs. representational efficiency}

The length of a program (the representation of the desired behavior) is defined in this context as the number of instructions stored in the main program and sub-processes. Note that this often differs from the number of actions that the program will \textit{generate}. As an illustration, the program in  Figure \ref{fig:lightbot} has length 13, but will generate a sequence of 38 actions. In calculating the length of a program, each action listed in a sub-process is counted once, and the entire sub-process sequence may then be executed for the cost of a single call in the main. An efficient program stores sequences of actions that are repeated multiple times as sub-processes, yielding a program that is shorter than the length of the flat sequence of actions it generates. With this task structure, we can design puzzles that tease apart the objectives of action efficiency and representational efficiency. We now turn to an empirical examination of human performance on the Lightbot task.

\section{Experiment: Exploring human abstractions}

\subsection{Methods}

\begin{figure}[ht]
\begin{center}
\includegraphics[width=.32\textwidth]{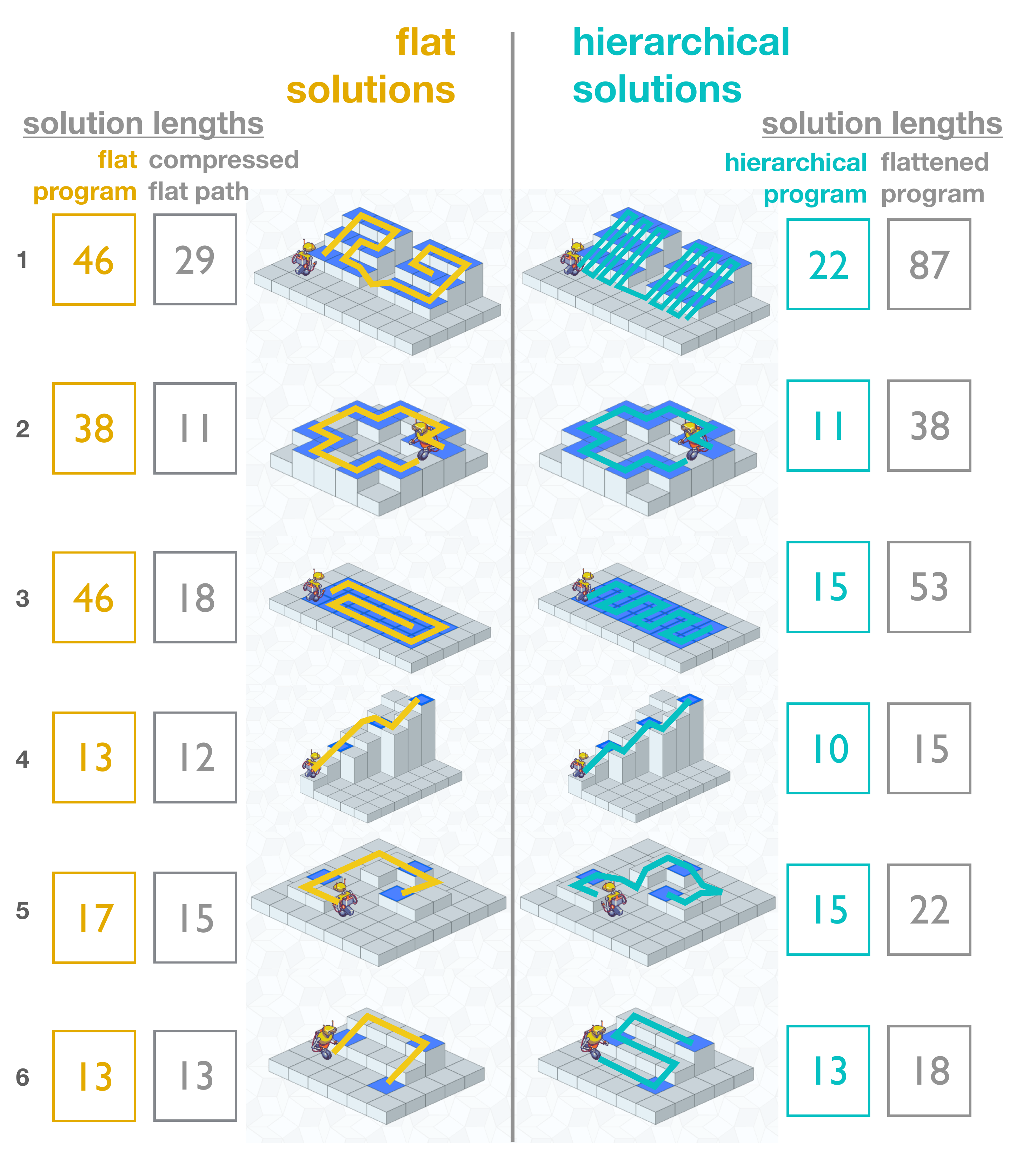}
\end{center}
\caption{Solutions to lightbot puzzles. Paths in yellow and blue show the execution traces of the best flat (yellow) and hierarchical (blue) programs. The lengths of the programs and their corresponding compressed (for flat) or flattened (for hierarchical) versions are displayed for each puzzle.}
\label{fig:solutions}
\end{figure}

We consider the six puzzles in Figure \ref{fig:solutions}, which illustrates the flat path that the optimal flat (left) and best hierarchical (right) solutions generate. All puzzles, with the exception of puzzle 2, are designed to tease apart action and representational efficiency. To demonstrate the distinction, consider the best flat and hierarchical solutions to puzzle 1. The shortest flat solution to this puzzle is 46 instructions long. Assuming 4 subprocesses can be stored, the most this solution can be compressed is into a hierarchical program of length 29. A considerably shorter hierarchical solution to this puzzle has length 22 in its compressed version. However, if one expands this program into the flat sequence of actions that it generates, the flat version is 87 instructions long. Thus, compression of the shortest flat solution is not a general strategy for obtaining the optimal hierarchical solution. 
 
For puzzles 1 and 3, the flat paths generated by the best hierarchical solutions are substantially longer than the shortest flat solutions. Puzzles 4, 5, and 6 were designed to meet the additional constraint that the shortest flat solution is incompressible or marginally compressible, meaning that there is little gain achieved by encoding the flat path into a hierarchical program. Puzzle 2 meets neither of these requirements, but possesses a clear recursive solution and is of interest in analyzing the efficiency of participants' programs. This design allows us to study the trade-off between action efficiency and representational efficiency by examining the length and compressibility of participants' flattened solutions. 

\subsubsection{Optimal flat solutions}
The flat solutions for these puzzles are found by training a reinforcement learning agent with Proximal Policy Optimization (PPO) \cite{schulman2017proximal} with a clipped objective.
The state representation contains four components: the direction the lightbot is facing, the height of the current square, the coordinates of the lightbot, and a binary array with an entry for each light, with a value of 1 if that particular light has been turned on, and 0 otherwise.
We encode the state with a binary vector representation as input to a two-layer neural network policy, which outputs a probability distribution over the actions. We also train a value network that estimates the value of each state. The agent receives -1 reward for each action taken and +1 reward for each light turned on. Once a light has been turned on, it cannot be turned back off. For each puzzle, we train the agent until convergence and then run 100 random rollouts from the trained policy. We use the rollout with the fewest number of actions as the standard for our flat solutions.

\subsubsection{Best hierarchical solutions}
A general method for obtaining optimal hierarchical solutions to Lightbot puzzles remains an open question. Thus, we consider our best approximations, which we obtained from the original creators of the task (puzzles 1, 2, and 3) or designed ourselves (puzzles 4, 5, and 6).

\subsubsection{Conditions}
In a between-subjects design, we assign participants to four conditions in which they solve the six test puzzles in variants of the Lightbot task. The four conditions differ as follows:
\begin{enumerate}[noitemsep]
    \item\underline{Efficient Flat}: Participants write flat solutions and are not given the capacity to use sub-processes. Participants are instructed to write solutions that use the fewest number of primitive actions and receive monetary bonuses proportional to their performance.
    \item\underline{Default Flat}: Participants write flat solutions and are not given the capacity to use sub-processes. Participants receive no instructions regarding solution length and receive a fixed bonus for each puzzle they complete.
    \item\underline{Efficient Hierarchy}: Participants are given the capacity to store four sub-processes per puzzle, which they may call any number of times. Participants are instructed to write programs that use the fewest number of stored instructions and receive bonuses proportional to performance. A counter displaying the length of the participant's program is visible while the participant generates solutions.
    \item\underline{Default Hierarchy}: Participants are given the capacity to store four sub-processes per puzzle, which they may call any number of times. Participants receive no instructions regarding program length and receive a fixed bonus for each puzzle they complete. No program length counter is present.
   \end{enumerate}
In all conditions, there are no imposed limits on the number of instructions in a sub-process or main program. 
  
\subsubsection{Task structure}
All participants first read through an illustrated tutorial explaining the game objectives and mechanics. In the Hierarchy conditions, this includes explanations of how to use sub-processes and how to call a sub-process within itself. In the Efficient Flat and Efficient Hierarchy conditions, participants additionally receive explanations and examples of action efficiency (Flat) or program efficiency (Hierarchy). 

Participants construct programs by dragging and dropping instructions into a program frame and are given the capacity to reorder and delete instructions. Participants can test their program at any time, which initiates an animation of Lightbot executing the program from the beginning. When a test of the program results in the completion of the level (all light tiles turned on), the experiment progresses to the next puzzle. Participants are given the capacity to skip a level after six minutes but receive no bonuses for skipped levels. Participants in all conditions receive the same set of nine puzzles: three simple tutorial puzzles presented in fixed order followed by the six test puzzles displayed in Figure \ref{fig:solutions}. Puzzles 1, 2, and 3 are presented first in random order, followed by puzzles 4, 5, and 6 in random order.

\subsubsection{Participants}
203 participants were recruited through Amazon Mechanical Turk and completed the Lightbot task in a web application. We excluded participants who reported experience with computer programming (39 participants) leaving a final participant pool of 164. No participants reported prior experience with the Lightbot game. The application was presented in full-screen mode to eliminate participant multitasking. The task required an average of 40 minutes across all conditions. Participants were assigned to one of four conditions: Efficient Flat, Default Flat, Efficient Hierarchy, and Default Hierarchy. Participants in the Efficient conditions were paid \$5.00 as a base rate and participants in the Default conditions received a \$4.50 base rate. Participants in all conditions were able to earn up to an additional \$3.00 in bonuses. In the Efficient Flat conditions, all participants who gave the shortest solution to a puzzle out of the participant pool received a \$0.50 bonus per puzzle. Longer solutions received bonuses that decreased linearly with their distance to the shortest solution. Participants were not informed the value of their bonuses during the task. In the Default conditions, participants received a bonus of \$0.50 for each test puzzle that they completed regardless of program efficiency. Participants received an average bonus of \$2.64 across all conditions. For all analyses, we include only data from complete solutions (not skipped). Thus, we are left with an average of 37 solutions per puzzle in the Efficient Flat condition, 38 in the Default Flat condition, 35 in the Efficient Hierarchy condition, and 46 in the Default Hierarchy condition. 

\section{Results and Analyses}
We first examine the efficiency of participants' flat solutions in the flat conditions to determine whether participants participants optimize action efficiency for flat solutions.

\subsection{Flat solutions}

\begin{figure}[ht]
\begin{center}
\includegraphics[width=.2\textwidth]{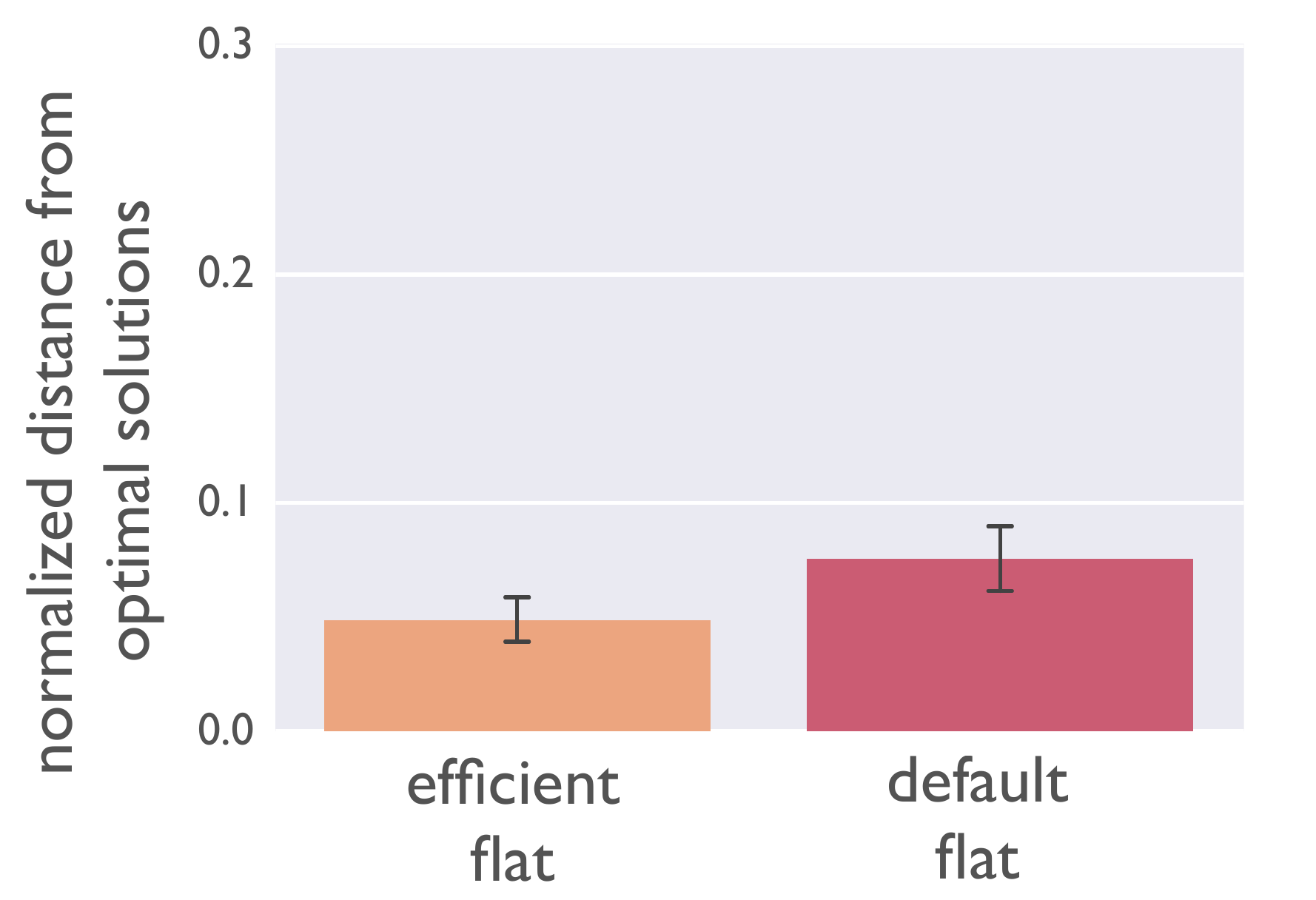}
\end{center}
\caption{Mean normalized distance of the length of participants' solutions in the flat conditions to the lengths of the optimal flat solutions, across all test puzzles. Error bars show 95\% CI.} 
\label{fig:flat_dist}
\end{figure}

Figure \ref{fig:flat_dist} shows the mean distance between the lengths of participants' flat solutions and the optimal flat solutions. Distance is in number of instructions and is normalized per puzzle by the min and max distances of the provided solutions, to facilitate comparison across puzzles. Across all puzzles, participants write solutions that are close but significantly longer than the optimal flat solutions in both the Efficient Flat ($p<.0001$) and Default Flat ($p<.0001$) conditions. There is a significant difference between the two conditions, with participants in the Default Flat condition generating solutions that are farther from the optimal solutions than participants in the Efficient Flat solutions ($t=2.98, p=.003$). Notably, the mode distance from optimal solutions in both of the flat conditions is $0$; $61\%$ of the solutions participants gave in the Efficient Flat condition and $55\%$ of the solutions participants gave in the Default Flat condition are optimal flat solutions. 

These results indicate that participants are highly effective at finding solutions that optimize action efficiency, even without explicit instruction.

\subsection{Hierarchical solutions}
\subsubsection{Length of flattened programs}
We now ask whether the hierarchical solutions given by participants optimize efficiency in action or representation. We address this question by analyzing the length and compressibility of the ``flattened'' version of participants' hierarchical programs. A flattened program consists of all actions that a hierarchical program generates. For example, the flattened program in Figure \ref{fig:lightbot} consists of 38 instructions, whereas the hierarchical program consists of 13. If participants optimize efficiency in the number of actions their program generates, we would expect to see no difference in the length of the flattened solutions given by participants in the Hierarchy condition and participants in the Flat conditions. Further, all of the puzzles we have used (with the exception of puzzle 2) are designed so that the flat version of the best algorithmic solution is {\em longer} than the best flat solution. Thus, if participants generate hierarchical solutions by minimizing program complexity, we expect to see longer flattened solutions to these puzzles. 

\begin{figure}[ht]
\begin{center}
\includegraphics[width=.35\textwidth]{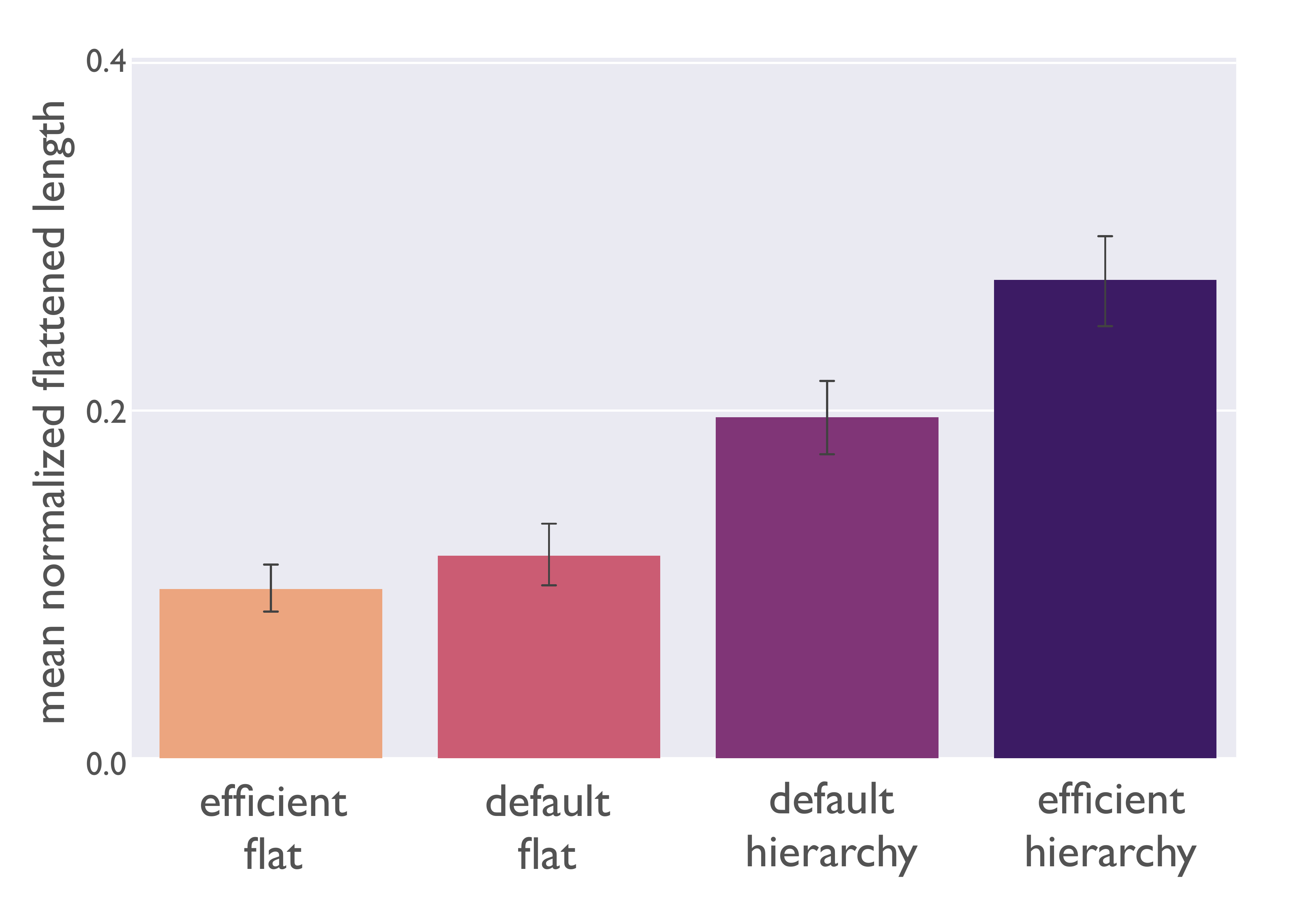}
\end{center}
\caption{Mean normalized flattened program length, across all test puzzles. Error bars show 95\% CI.} 
\label{fig:normalized_flat_program_length}
\end{figure}

Figure \ref{fig:normalized_flat_program_length} shows the mean lengths of the flattened versions of programs participants generate, normalized per puzzle by the min and max flat lengths. Aggregating across all puzzles, the mean normalized flattened length in the Default and Efficient Hierarchy conditions is significantly greater than in the Efficient Flat (Default Hierarchy: $t=6.55, p<.0001$; Efficient hierarchy: $t=10.334, p<.0001$) and Default Flat conditions (Default Hierarchy: $t=4.00, p<.0001$; Efficient Hierarchy: $t=7.85, p<.0001$). The {\em programs} that participants write in the Hierarchical conditions are, however, significantly shorter than the flat solutions provided in the Efficient (Default Hierarchy: $t=3.47, p<.001$; Efficient Hierarchy: $t=6.49, p<.0001$) and Default Flat conditions (Default Hierarchy: $t=3.76, p<.001$; Efficient Hierarchy: $t=6.74, p<.0001$). 

This dissociation reflects the trade-off between action and representational efficiency built in to the solutions and suggests that participants do not simply optimize action efficiency when generating hierarchical solutions. Further, the flattened versions of solutions participants generate in the Efficient Hierarchy condition are also significantly longer than those generated in the Default Hierarchy condition ($t=4.60, p<.0001$), despite programs in the Efficient Hierarchy condition being significantly shorter than those in the Default Hierarchy condition ($t=5.11, p<.0001$), showing that this dissociation becomes stronger when participants are explicitly instructed to optimize representational efficiency.

To more explicitly target the representational efficiency of participants' solutions, we analyze the explicit ``compressibility'' of participants' hierarchical and flat solutions. 

\subsubsection{Compressibility of flat paths}
Puzzles 4, 5, and 6 were designed so that the shortest flat solutions is only marginally compressible, whereas the path that the best hierarchical solution generates contains considerable repeated structure that can be exploited by a hierarchical program. Thus, participants who optimize action efficiency will produce solutions to these puzzles whose flattened versions will show minimal compressibility.

To quantify the degree of compressibility in participants' solutions, we implement a simple algorithm that compresses a flat sequence of actions into a hierarchical program. The algorithm iteratively stores the longest, most-repeated sub-sequence as a subprocess until there are no more repeated subsequences it can compress, or it has stored 4 subprocesses. We impose the minimal requirement that the sub-sequence must contain at least two actions and must be repeated at least twice to be stored as a sub-process. After compressing a sequence, the algorithm checks whether some recursion is possible, by identifying whether the main program consists only of a repeated call to a single subprocess. If it does, the algorithm appends the sub-process label to the end of that sub-process and replaces the main program with a single call to the sub-process.

Given a flat path of actions, this algorithm is able to generate compact, hierarchically structured programs that are close approximations to optimal programmatic compressions of flat paths. We use this algorithm to compress the flat and flattened solutions generated by participants to compute the compressibility of a participant's solution as:$$compressibility = \frac{\textit{flat length} - \textit{compressed length}}{\textit{flat length}}$$ \begin{figure}[ht]
\begin{center}
\includegraphics[width=.35\textwidth]{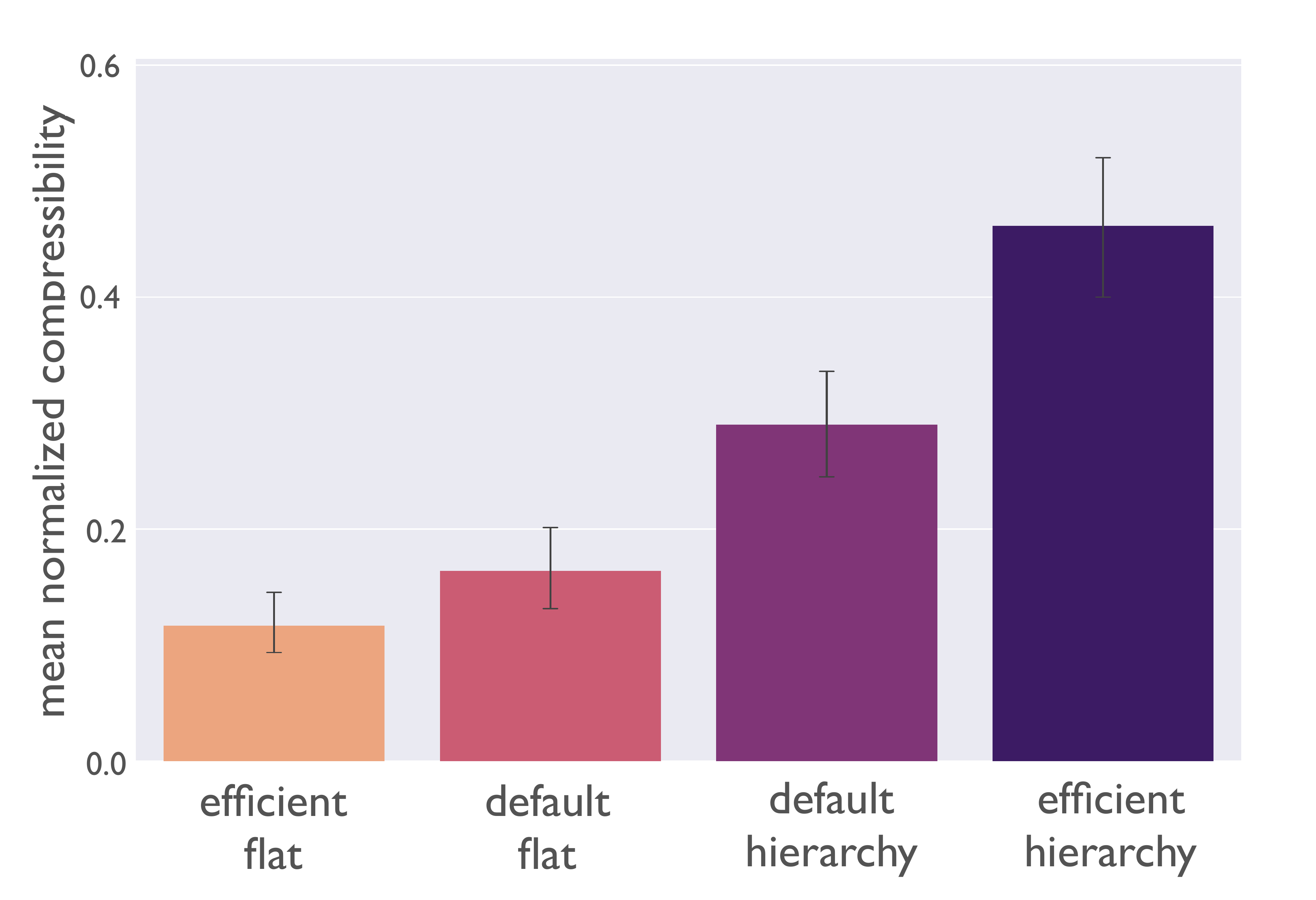}
\end{center}
\caption{Mean normalized compressibility for puzzles 4, 5 and 6. Error bars show 95\% CI.} 
\label{fig:compressibility_by_puzzle}
\end{figure}

Figure \ref{fig:compressibility_by_puzzle} shows rates of compressibility for the three relevant puzzles, normalized per puzzle by the min and max score. Across all three puzzles, participants in the Efficient Hierarchy and Default Hierarchy condition produce solutions whose flattened versions are more compressible than those in the Efficient (Default Hierarchy: $t=6.51, p<.0001$; Efficient Hierarchy: $t=8.50, p<.0001$) and Default Flat (Default Hierarchy: $t=4.80, p<.0001$; Efficient Hierarchy: $t=9.11, p<.0001$) conditions. Taken with the finding that participants in the Hierarchy conditions generate solutions are longer in the actions that they generate, this finding supports the hypothesis that participants are maximizing representational efficiency over action efficiency. 

We also see a significant difference in the compressibility of solutions generated in the Efficient and Default Flat conditions for puzzles 5 and 6 (5: $t=2.05, p<0.05$; 6: $t=2.22, p<.05$; no significance for puzzle 4: $t=.3, p=.76$), for which participants in the Default Flat condition generate solutions that are significantly more compressible than those generated in the Efficient Flat condition. This is despite the fact that participants are writing flat solutions (for which compressibility confers no explicit advantage), as well as the fact that the compressible solutions are longer than the incompressible, efficient solutions to the problem. Participants in the Default Flat condition received minimal guidance for how to solve Lightbot problems, so this condition provides insight into the `default' method for solving these problems. Although there is no explicit advantage to using compressible flat sequences in the Default Flat condition, there is an implicit advantage in reducing working memory demands while planning. This exemplifies the idea that minimizing representational complexity may be a natural, default objective for a problem solving task with sufficient algorithmic complexity.

\section{Discussion}

The hierarchical reinforcement learning literature has focused on learning hierarchical abstractions that allow an agent to efficiently navigate the problem space, with less emphasis on modeling the \textit{structure} of the specific problem currently under consideration. Indeed, recent work has emphasized learning to chain together temporally-extended actions, such as learning high level controllers that learn to break tasks down into subgoals \cite{vezhnevets2017feudal,kulkarni2016hierarchical}, learning options and their termination conditions simultaneously \cite{bacon2017option,harb2017waiting}, and learning reusable behaviors shared across tasks \cite{frans2017meta,florensa2017stochastic}. However, a problem may exhibit much more algorithmic structure than a linear chain, highlighting the large gap between the limited expressiveness of current HRL methods and the rich structure of problems that demand abstraction.

Here, we have introduced a task compatible with the RL framework that introduces a novel objective: solving the task as \textit{computationally efficiently} as possible. With this task, we find that naive participants show a preference for generating hierarchical solutions that optimize representational efficiency over action efficiency, which emphasizes the importance of representation learning and efficient coding for reinforcement learning. In future empirical work, we plan to examine other structural properties of the hierarchical solutions participants generate that may trade off with program length, such as symmetry, balance, depth, and the branching factors of their execution trees. Our task facilitates conversation between hierarchical approaches to problem solving in cognitive science and computer science, and may be used as a benchmark task in future computational work.

\vspace{2mm}

\begin{small}
\noindent {\bf Acknowledgments.} This work was supported by Air Force Office of Scientific Research grant number FA9550-18-1-0077.
\end{small}

\renewcommand{\bibliographytypesize}{\small}
\bibliographystyle{apacite}
\setlength{\bibleftmargin}{.125in}
\setlength{\bibindent}{-\bibleftmargin}

\bibliography{bib}

\end{document}